\documentclass{article}
\usepackage{spconf,amsmath,graphicx}
\usepackage{amssymb}
\usepackage{soul}
\usepackage{color}
\usepackage{enumitem}

\usepackage[colorlinks = true,
            linkcolor = blue,
            urlcolor  = blue,
            citecolor = blue,
            anchorcolor = blue]{hyperref}

\usepackage{caption}
\usepackage{subcaption}
\usepackage{environ}
\NewEnviron{myequation}{%
\begin{equation}
\scalebox{1.0}{$\BODY$}
\end{equation}
}

\usepackage{xcolor}
\usepackage[linesnumbered,ruled,vlined]{algorithm2e}

\SetCommentSty{mycommfont}

\SetKwInput{KwInput}{Input}                
\SetKwInput{KwOutput}{Output}              

\usepackage[font=small,skip=3pt]{caption}

\title{Level set based particle filter driven by optical flow: an application to track the salt boundary from X-ray CT time-series} 

\name{Karim Makki$^{a,*}$\thanks{$^*$These authors contributed equally.}, Jean François Lecomte$^{a,*}$, Lukas Fuchs$^{c}$, Sylvie Schueller$^{a}$, Étienne Mémin$^{b}$}
\address{$^{a}$ IFP Energies nouvelles, 1-4 avenue de Bois-Préau, 92852 Rueil Malmaison, France \\ $^{b}$ INRIA Rennes Atlanique, IRISA, France \\ $^{c}$ Institute for Geosciences, AG Geodynamic, Goethe University Frankfurt, Germany}

\begin{document}
\maketitle
\begin{abstract}

Image-based computational fluid dynamics have long played an important role in leveraging knowledge and understanding of several physical phenomena. In particular, probabilistic computational methods have opened the way to modelling the complex dynamics of systems in purely random turbulent motion.
In the field of structural geology, a better understanding of the deformation and stress state both within the salt and the surrounding rocks is of great interest to characterize all kinds of subsurface long-terms energy-storage systems. 
The objective of this research is to determine the non-linear deformation of the salt boundary over time using a parallelized, stochastic filtering approach from x-ray computed tomography (CT) image time series depicting the evolution of salt structures triggered by gravity and under differential loading.  
This work represents a first step towards bringing together physical modeling and advanced stochastic image processing methods where model uncertainty is taken into account.
\end{abstract}
\section{Introduction}
\label{sec:intro}
Tracking the contours and the motion of highly deformable structures is an essential task in many computer vision applications. In particular, following the interface delineating between iso-quantities transported by a fluid flow through image sequences remains a difficult task. This particular case is of importance in domains such as structural geology where one may wish to track the salt/rock interface buried in sedimentary basins. In fact, the presence of salt in a basin affects virtually all aspects of the structural system~\cite{baumann2018deformation}. Therefore, a better understanding of the deformation and stress state both within the salt and the surrounding rocks is crucial to characterize all kinds of subsurface long-terms storage systems.
Generally, rock salt buried in sedimentary basins can be considered to behave like a viscous fluid.
To resolve transport and topological changes, a natural option is to control the system state in terms of an advection equation, in particular a Hamilton–Jacobi equation. 
The most popular deterministic methods for advective flow problems can be classified into two main categories: non-parametric (optical flow~\cite{horn1981determining}, Level set~\cite{osher1988fronts}), and parametric (the large deformation diffeomorphic metric mapping (LDDMM)~\cite{beg2005computing}, and its derivative metamorphosis model~\cite{trouve2005metamorphoses, maillard2022deep}). 
Although parametric methods offer a standard approach for establishing \textit{one-to-one} point correspondences, fluid flow maps are not necessarily bijective. Alternatively, we use a non-parametric formulation to advect an implicit interface while introducing a second vectorial level set to determine and maintain point correspondences across time~\cite{pons2006maintaining}. This choice will enable exploitation of all the available photometric information, thus avoiding extraction of meaningful control points, which are difficult to define due to non-linear interference in image intensities.
Unfortunately, deterministic advective models alone are insufficient to completely represent the random nature of fluid motion and  to deal effectively 
with some critical situations ~\cite{cotter2020particle,cintolesi2020stochastic}. For example, in the case of missing local photometric information related to time scale and resolution issues. Indeed, the tracked structure may get severely distorted, and the evolving interface may vanish over several time steps.
To move from a deterministic to the modelling paradigm, \textit{Avenel et al.} proposed a stochastic‐advective transport model for the tracking of non-convex closed curves through image data~\cite{avenel2014stochastic}. Contrary to approaches relying on exact physical evolution models, it considers a simplified advection dynamics while introducing additional uncertainty terms to further predict the most \textit{probable} curve trajectory
among a considerable number of \textit{turbulent} trajectories using a Kalman-like filter. This method was originally dedicated to track curves whose appearance is not known in advance. It has been successfully applied to various satellite image sequences portraying the evolution of complex geophysical flows. It is therefore a promising tool to explore salt dynamics under the assumption that the predominant form of transport of the salt/rock interface should be advective.
In this context, and since we dispose of both current and future state observations, we propose to guide the random curve’s samples toward meaningful areas of the state space by optical flow velocity. Although a similar approach has been used in human motion tracking~\cite{tung2008human}, this article represents an attempt to highlight its capacity to solve slightly more difficult problems such as the tracking of closed curves transported by fluid flows using the level set formalism. The method is finally applied to X-ray CT fluid flow images within deformed rock salt.

\section{Material and Methods}
\label{sec:mm}

\subsection{Temporal data sequences}
\label{data_acquisition}

Image-based computational fluid dynamics can help to establish and validate a stable and accurate model to examine the kinematics of salt flow during the evolution of a salt structure triggered by gravity along a slope and under differential loading. In this work, following the protocols used in~\cite{callot2016three}, laboratory scaled (analogue) models were performed, allowing to mimic the kinematics and structural evolution of complex systems, where salt structures are involved. Salt is modelled by silicone putty, and the sediments are represented by layers of dry sand or glass microbeads. Computerized X-ray tomography is applied to these models in order to acquire the kinematic evolution and the 3D geometry without interrupting or destroying the model~\cite{colletta1991computerized}. Scanner data used for this work corresponds to collections of 2D images repeatedly acquired in time along different parallel sections within the models.

\subsection{Image processing methods}

\subsubsection{Preliminaries}

The optical flow method is a classical technique for determining flow velocity from image brightness variations. The connection between optical flow and fluid flow is explored in the studies~\cite{liu2008fluid,heitz2010variational}, providing insights into its application to image-based fluid velocity measurements. 
Since its discovery by Horn and Schunck in~\cite{horn1981determining}, several enhancements of the original formulation were proposed by introducing different smoothness constraints and penalty terms to the energy functional to be minimized~\cite{nagel1986investigation}.
Instead of focusing on regularization terms, another strategy was employed in~\cite{tauro2018optical} by using the original formulation, followed by a posteriori filtering to get a realistic velocity flow. The idea behind was to only retain velocity exhibiting minimum variance of length and angle of travel between successive images in a given sequence. 
To perform robust tracking and prevent drift effects, a more generic particle filter has been originally introduced in~\cite{avenel2014stochastic}  where the drift component was determined through the projection of a deterministic transport velocity $\omega$ on curve's normal $n$ (to evolve the curve by front propagation according to the level set principle), combined with a supplementary data-driven force $F$ estimated using the Chan Vese method~\cite{vese2002multiphase} according to the equation:
$\omega_n = \beta n^T \omega
+(1-\beta)\partial_{\varphi} F(\varphi)$, for $\beta \in [0,1]$ and given a dynamic signed distance function $\varphi$ representing the tracked interface implicitly.

\subsubsection{Proposed approach}
In this paper we combined the optical flow method~\cite{horn1981determining} with a stochastic filtering method for the tracking of level sets~\cite{avenel2014stochastic} with the aim to track closed curves delineating between iso-quantities transported by fluid flows.
Particle filter, also known as the sequential Monte Carlo method, is a technique to solve hidden Markov models and nonlinear (non Gaussian) models. 
It allows for modelling the evolution of a dynamical system according to a stochastic differential equation (SDE):

{\scriptsize
\begin{equation}
    \begin{cases}
      dx(t) = f(x,t) dt + \sigma_t d\textbf{B}(x,t)\\
      z_k = g(x_k)+v_k
      
    \end{cases}\,,
    \label{SDE}
\end{equation}}\normalsize
where $x(t)$ is the system state at time $t$, $f$ is a deterministic evolution function, $\sigma_t$ is a diffusion constant, $\textbf{B}$ is a Brownian motion, $z_{1:k}= \{z_1 \ldots z_k\}$ is a discrete set of observations, and $v$ is the error in measures (\textit{e.g.} image noise). \\
The method relies on estimating the transition probability of system state according to the following pdf:

{\scriptsize
\begin{equation}
\pi(x_k|x_{k-1}) \sim \mathcal{N}(\underbrace{x_{k-1}+f(x_{k-1},t_{k-1}).(t_k-t_{k-1})}_{\mu_k}, \sigma_{k} )
    \label{transition}
\end{equation}}\normalsize

The above defined approximation can be interpreted as a first-order Euler approximation to the deterministic evolution $\mu_k$, convolved with a distribution of perturbations $\mathcal{N}(0, \sigma_{k})$. Equivalently, the
space of plausible turbulent solutions is centered at the deterministic solution $\mu_k$.  The \textbf{predicted} system state is therefore given by the probability distribution defined in Eq~\ref{transition}. Subsequently, a correction step \textbf{updates} the posterior pdf
through the likelihood $\pi (z_k|x_k)$ based on the new observation $z_k$.
The different steps of the algorithm used in this work, namely the bootstrap particle filter, are detailed in Algorithm~\ref{bootstrap}. In this work, the system state is represented as a signed distance function $\varphi$ defined in the image domain $\Omega$ (negative inside salt structure and positive outside). Being a closed curve, the salt/rock interface $\Gamma$ was implicitly represented as the zero level set of $\varphi(\textbf{x},t): \Omega \times \mathbb{R}_+ \mapsto \mathbb{R}$, such that $\Gamma = \{\varphi = 0\}$.
The proposed model relies on creating an uncertainty on interface's motion along tangent (to take into account the viscous flow behaviour) and normal (to model pressure effect) directions. 
The underlying control SDE reads:

{\scriptsize
\begin{equation}
\boldsymbol{d\varphi}(t) = 
\underbrace{
\omega_n
\boldsymbol{n}(t)
dt}_{\text{deterministic term}} +
\underbrace{
\begin{bmatrix}
\boldsymbol{n}(t)\\
\boldsymbol{n_{\perp}}(t)
\end{bmatrix}^{tr}
\begin{bmatrix}
\sigma_n & 0\\
0 & \sigma_{\perp}
\end{bmatrix}
\begin{bmatrix}
d\mathbf{B_n}(t) \\
d\mathbf{B}_{\perp}(t) 
\end{bmatrix}}_{\text{probabilistic term}}
\label{eqn1}
\end{equation}}\normalsize

where: $n = \frac{\nabla \varphi}{|\nabla \varphi|}$ is the outward normal to the interface, $n_{\perp} = \frac{1}{|\nabla \varphi|}(\partial_y \varphi, -\partial_x \varphi)^T$ is the unit vector tangent to the interface. $\sigma_n$ and $\sigma_{\perp}$ are two diffusion constants, $\mathbf{B_n}$ and $\mathbf{B}_{\perp}$ are two uncorrelated Brownian motions, and $\omega_n = \omega \cdot n$ is the projection on the curve’s normal of a deterministic transport velocity field $\omega$. For instance, this velocity flow can be an optical flow.
Let us recall that the optical flow method relies on advecting image intensities $I$ according to a velocity flow $\omega$ satisfying: 

{\scriptsize
\begin{equation}
\partial_t I(\textbf{x},t) + \omega.\nabla I(\textbf{x},t)=0.
\label{intensity_advection}
\end{equation}}\normalsize
 
Assuming that $\varphi(\textbf{x},t)$ is a function of both position $\textbf{x} = (x, y)$, and time $t$, 
one can infer directly the curve’s velocity field from each particle displacement by introducing a second vectorial level set $\psi(\textbf{x},t)$ to determine point correspondences between successive images. In the deterministic case (\textit{i.e.} $\sigma_n = \sigma_{\perp}= 0$), both $\varphi$ and $\psi$ are advected according to a certain velocity field $\omega$:

{\scriptsize
\begin{equation}
    \begin{cases}
      \partial_t \varphi(\textbf{x},t) + \omega . \nabla\varphi (\textbf{x},t) =0\\
      \partial_t \psi(\textbf{x},t) + D \psi(\textbf{x},t) \quad \omega =0
      
    \end{cases}\,,
    \label{levelset_advection}
\end{equation}}\normalsize

with initial data: $\varphi(\textbf{x},0) = \varphi_0$, $\psi(\textbf{x},0) = \psi_0$, and where $D \psi$ is the Jacobian matrix of $\psi$. 

Since $I$, $\varphi$ and $\psi$ obey the same evolution law (see Eqns~\eqref{intensity_advection} and~\eqref{levelset_advection}), the deterministic transport velocity field from an image $k-1$ to its successor $k$ is physically predicted by the extended advection equation~\eqref{intensity_advection} for image intensities for varying optical flow over time. It is therefore obtained by minimizing a global energy functional:

{\scriptsize
\begin{equation}
    \omega^* =  \underset{(\omega_x, \omega_y)}{\textit{min}}\, \int \int [(I_x \omega_x + I_y \omega_y + I_t)^2 + \alpha ^2 (|\nabla \omega_x|^2 + |\nabla \omega_y|^2 ) ] dx dy,
    \label{energy}
\end{equation}}\normalsize

where $\alpha$ is a smoothness/regularization constant, $I_x$, $I_y$, and $I_t$ are the derivatives of the image intensity values along the $x$, $y$ and time dimensions respectively. 
The optical flow velocity is further applied to simultaneously make $\varphi$ and $\psi$ evolve according to~\eqref{levelset_advection} within the stochastic formalism.

{\scriptsize
\begin{algorithm}[!h]
\DontPrintSemicolon

  \textbf{Initialisation:} $t=0$
  
  \For{$i= 1,\ldots,N$}    
        { 
        	sample $\varphi_0 ^{(i)} \sim \pi(\varphi_0)$ and set $t=1$.
        }
\textbf{step 2: importance sampling} 

  \For{$i= 1,\ldots,N$}    
        { 
        \begin{itemize}[noitemsep,topsep=0pt]
            \item 
        	sample $\tilde{\varphi}_t ^{(i)} \sim \pi(\varphi_t / \varphi_{t-1}^{(i)})$ based on the \\ Markov process model and set \\ $\tilde{\varphi}_{0:t}^{(i)} = (\varphi_{0:t-1}^{(i)}, \tilde{\varphi}_t^{(i)}) $.
        	\item 
        	evaluate the importance weights (reweight):\\ $\tilde{W}_t^{(i)} = \pi(z_t|\tilde{\varphi}_t^{(i)})$.

        	\item
        	normalise the importance weights.
        	 \end{itemize}
        }

\textbf{Step 3: selection} 

Resample with replacement $N$ particles $\{\varphi_{0:t}^{(i)}\}_{i=1,\ldots,N}$ from the set $\{\tilde{\varphi}_{0:t}^{(i)}\}_{i=1,\ldots,N}$ according to the importance weights.

Set $t \leftarrow t+1$
    
\If   { $t \leq T$} 
    {
            go back to step 2.
     }
\Else{ exit.}
\caption{bootstrap particle filter}
\label{bootstrap}
\end{algorithm}}\normalsize

\section{Experiments and results}

\subsection{Model evaluation}
\begin{figure}[h!]
\begin{minipage}[b]{.48\linewidth}
  \centering
  \centerline{\includegraphics[width=4.0cm]{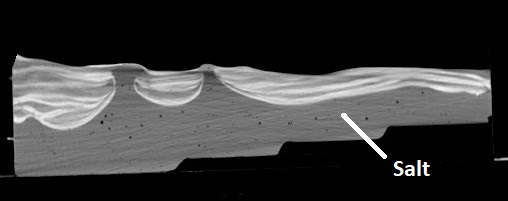}}
  \centerline{(a) CT scan}\medskip
\end{minipage}
\hfill
\begin{minipage}[b]{0.45\linewidth}
  \centering
  \centerline{\includegraphics[width=3.9cm]{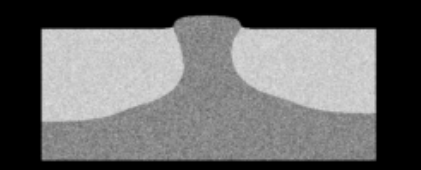}}
  \centerline{(b) Simulated}\medskip
\end{minipage}
\begin{minipage}[b]{.45\linewidth}
  \centering
  \centerline{\includegraphics[width=4.0cm]{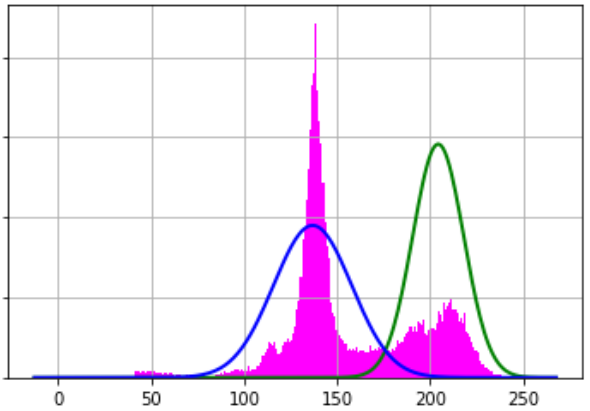}}
  \centerline{(c) Histogram of (a)}\medskip
\end{minipage}
\hfill
\begin{minipage}[b]{0.45\linewidth}
  \centering
  \centerline{\includegraphics[width=4.0cm]{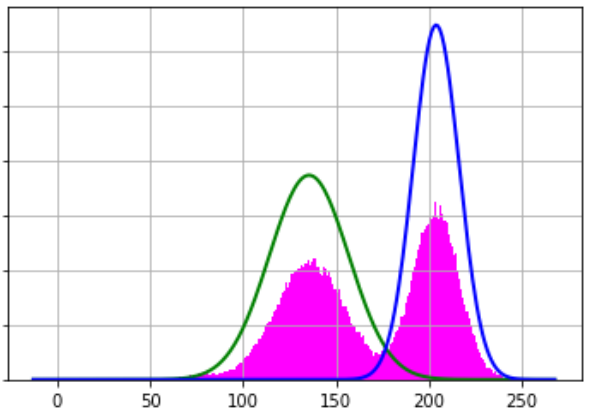}}
  \centerline{(d) Histogram of (b)}\medskip
\end{minipage}
\caption{Reproducing intensity dynamics of CT data.}
\label{fig:gmm}
\end{figure}

To evaluate the accuracy of our stochastic model in the context of salt boundary tracking, and since ground truth segmentations are difficult to obtain from CT data, we used a data sequence generated using the numerical model of salt diapir formation by down-building~\cite{fuchs2011numerical}. Initially, this sequence consisted of a set of temporal binary segmentations for three classes: salt layer, sediments, and image background. 
Since our approach is mainly intensity based, reproducing intensity dynamics nearly similar to those of realistic scanner images is crucial to the conduct of a valid accuracy assessment from synthetic data. To this end, we approximated a Gaussian mixture model to determine a sparse set of parameters (mean $\mu_i$ and variance $\sigma_i$ of each class $i$ from the histogram of a CT scan showing clearly the different structures). Note that, for a better classification, the background was ignored at this stage. Subsequently, we modeled the intensity distribution of each class with a normal distribution $\mathcal{N}_i(\mu_i,\sigma_i)_{i \in \{1, 2\}}$ (white noise) that provides the best fit to each class intensities.
A per-class (local) Gaussian filtering of standard deviation $\gamma = 2$ is further applied to have more strongly correlated noise without blurring the inter-class edges. An example for reproducing CT-like intensity dynamics is illustrated in Fig~\ref{fig:gmm}.

\begin{figure}[!htb]
\centering
\includegraphics[width=0.45\textwidth]{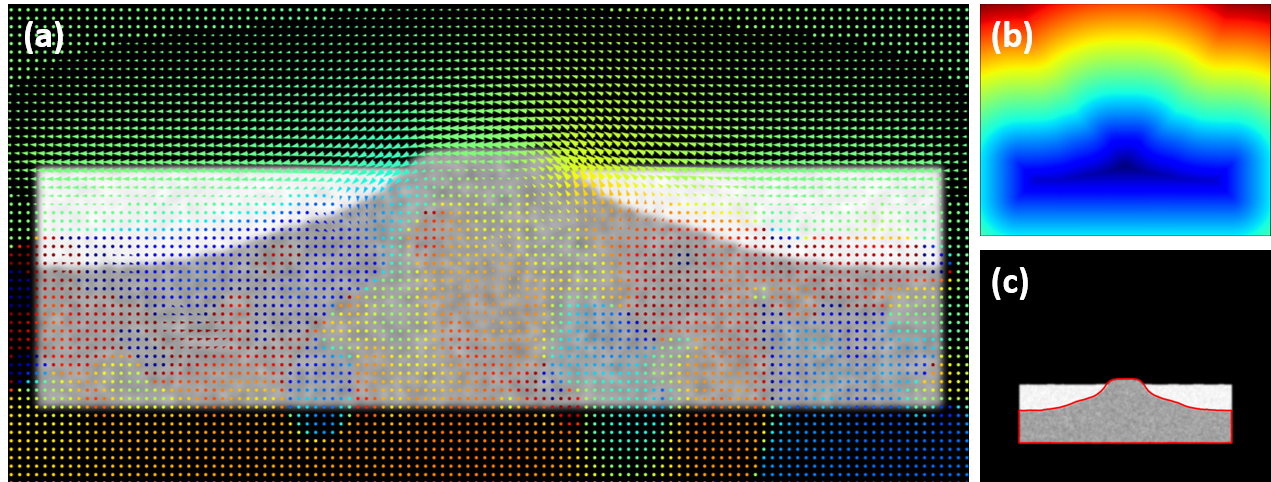}
\caption{\label{fig:evolving} Advecting the level set function according to the optical flow velocity: (a) estimated optical flow $\omega_{k-1 \rightarrow k}$, (b) $\varphi_k = \omega \circ \varphi_{k-1}$, and (c) current image $I_k$ on which the isocurve $\varphi_k = 0$ is plotted in red.}
\end{figure}

\begin{figure}[h!]
     \centering
     \begin{subfigure}[b]{0.11\textwidth}
         \centering
         \includegraphics[width=\textwidth]{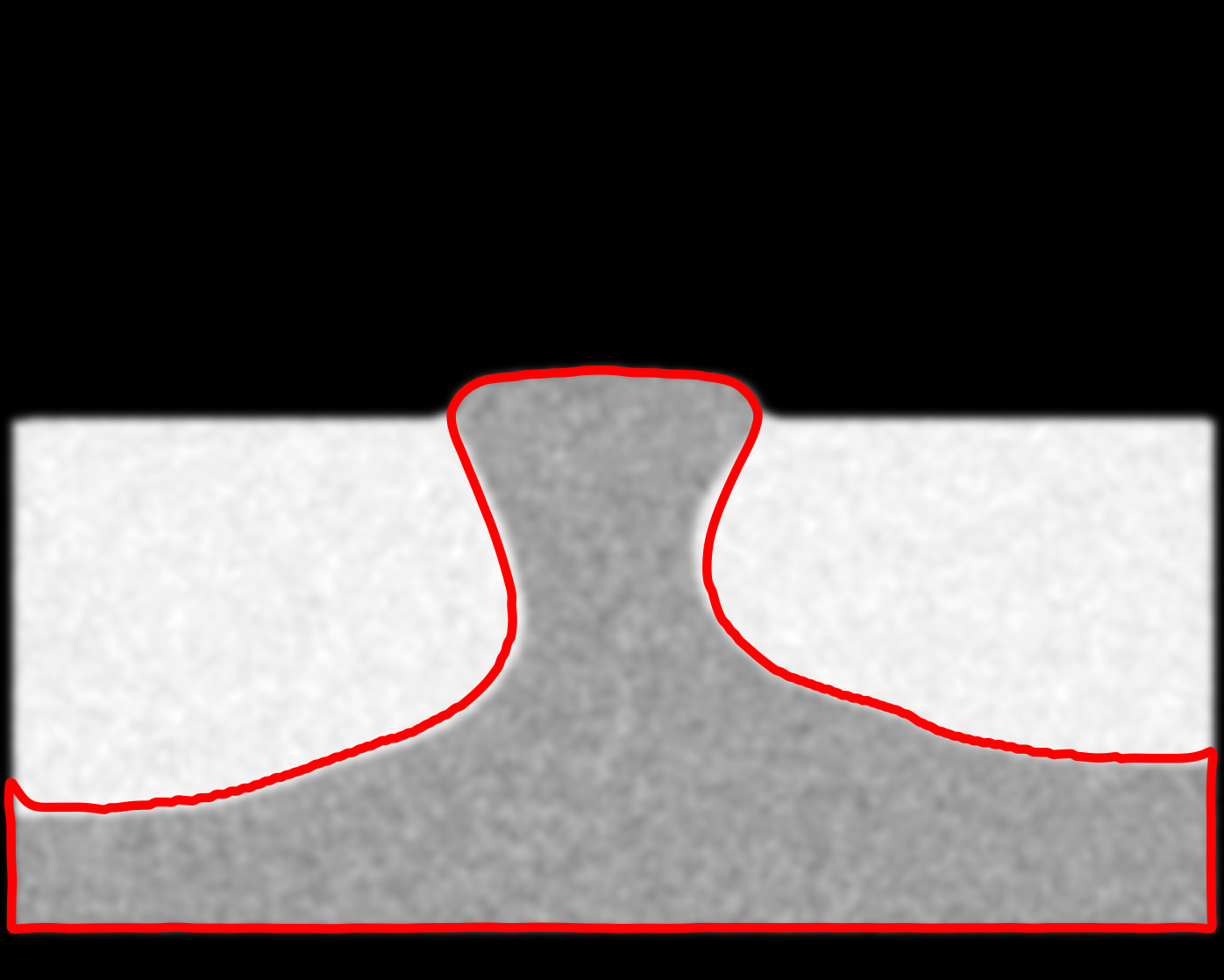}
         \caption{\#34}
     \end{subfigure}
     \begin{subfigure}[b]{0.115\textwidth}
         \centering
         \includegraphics[width=\textwidth]{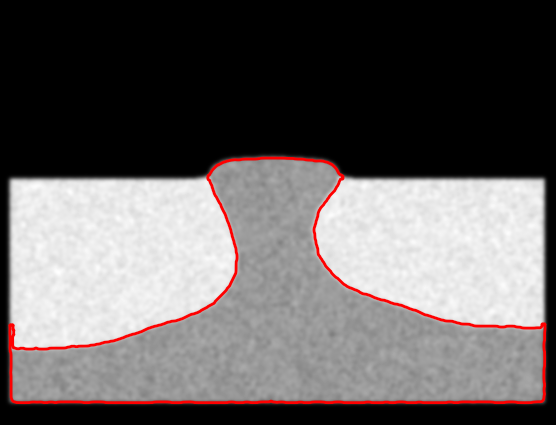}
         \caption{\#34}
     \end{subfigure}
          \begin{subfigure}[b]{0.11\textwidth}
         \centering
         \includegraphics[width=\textwidth]{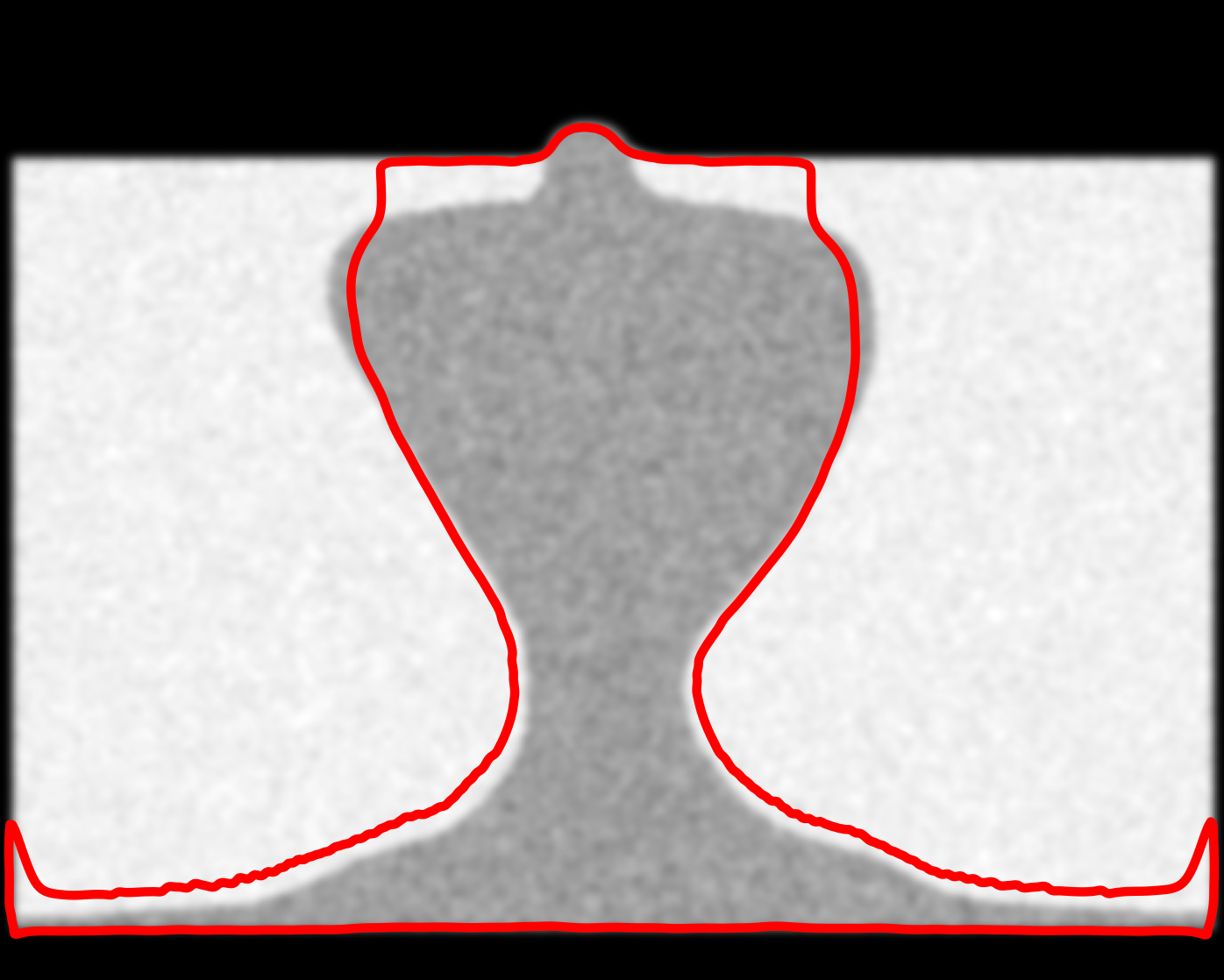}
         \caption{\#60}
     \end{subfigure}
     \begin{subfigure}[b]{0.113\textwidth}
         \centering
         \includegraphics[width=\textwidth]{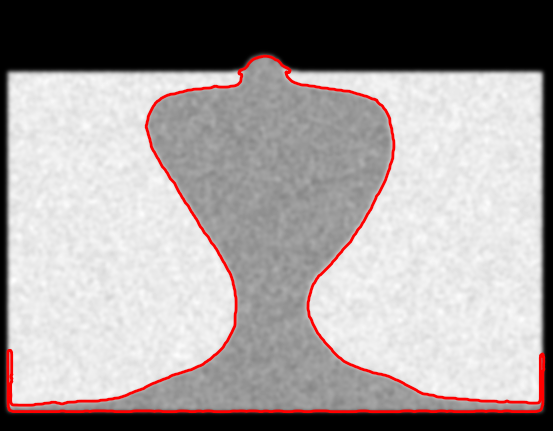}
         \caption{\#60}
     \end{subfigure}
    
        \caption{\label{fig:comparison} Deterministic Vs stochastic tracking. (a) and (c): deterministic. (b) and (d): stochastic approach. Sub-captions indicate time. The total accumulated RMSEs were $7.93$ and $1.34$ for the deterministic and stochastic propagative models, respectively.}
        \label{fig:realistic}
\end{figure}

\begin{figure}[!htb]
\centering
\includegraphics[width=0.3\textwidth]{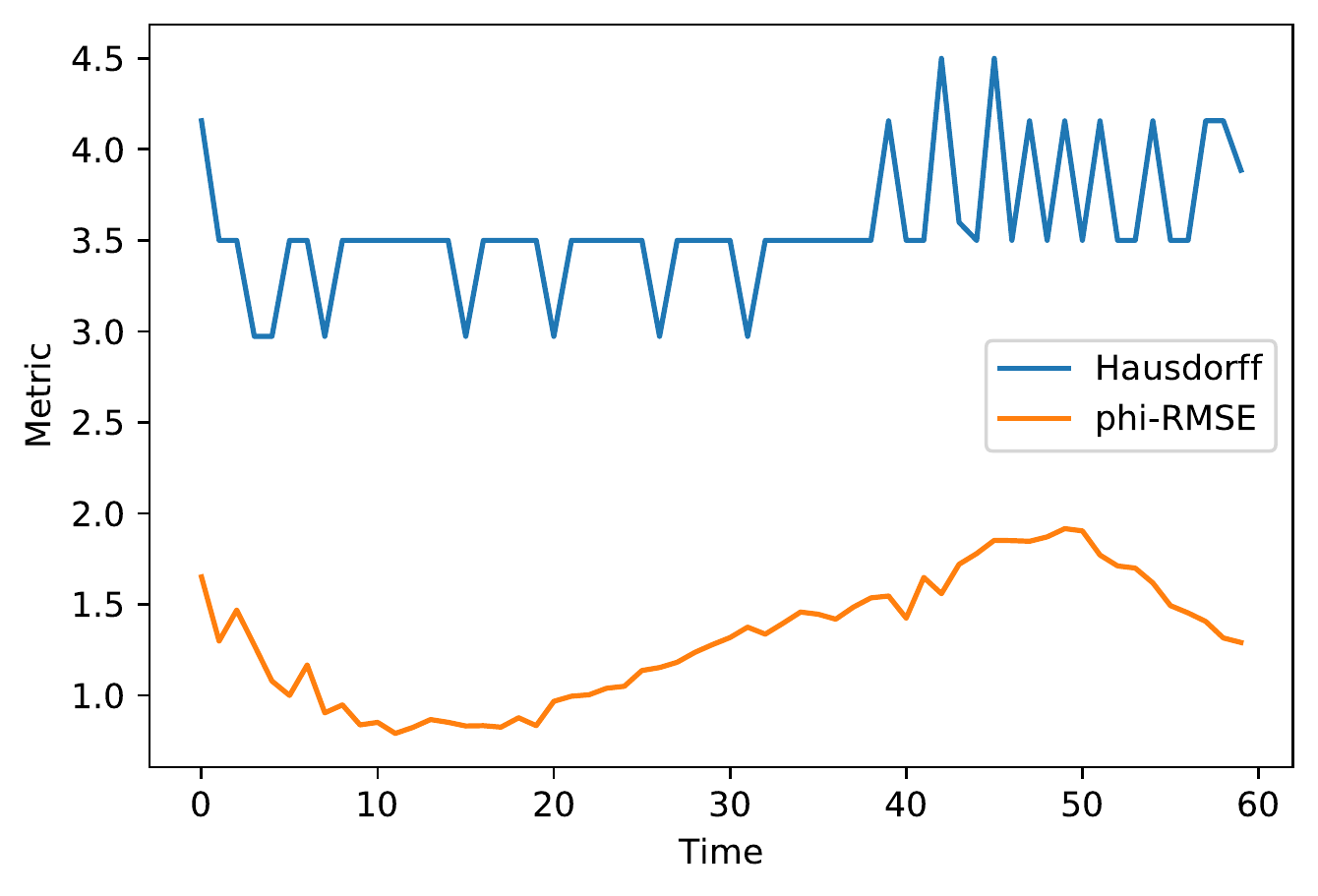}
\caption{\label{validation} Quantitative evaluation of segmentation accuracy.}
\end{figure}

Our intensity-based model was therefore evaluated by using this simulated sequence. As we use a continuous time propagative model, the objective of this step was two-fold: 1) to evaluate the sensitivity of both deterministic and stochastic models to error accumulation during tracking, 2) to quantitatively evaluate the accuracy of the employed stochastic model.   
To pilot the particle filter at each time, a smooth deterministic velocity field was first estimated between successive images by minimizing the energy functional of Eq~\eqref{energy} with $\alpha =7$ (see Fig.~\ref{fig:evolving}).
In the stochastic setting, 200 particles were generated, $\sigma_n$ and $\sigma_{\perp}$ were both set to 2 in Eq~\eqref{eqn1}.
Fig~\ref{fig:comparison} shows that the stochastic approach is less sensitive to the problem of tracking error propagation than the deterministic one. This may be explained by the permanent observation-based trajectory refinement and the subdivision of time interval between successive frames into 20 small steps.
As shown in Fig~\ref{validation}, two error metrics were used to assess the accuracy of the automatic segmentations (we refer to an automatic segmentation as the set $\mathcal{S}=\{\textbf{x} \in \Omega  |  \varphi(\textbf{x}) <= 0 \}$). In addition to the Hausdorff distance, we evaluated the temporal RMS error on the estimated scalar functions $\varphi$ in a narrow-band of $\pm 3$ pixels (px) around the 0-interface of their true homologous computed from ground-truth segmentations using the fast marching method.
These two metrics allowed the evaluation of maximal and average tracking errors, respectively.
A respectful level of tracking accuracy was achieved since we obtained a maximal distance between automatic and ground truth segmentations of the order of 4.5px and a mean error below 2. Errors tended to occur almost exclusively in the bottom corners. 

\subsection{Application to x-ray CT time-series}

As illustrated in Fig.~\ref{fig:realistic}, we were able to both track salt boundary, as well as the turbulent trajectory of a set of two curve points throughout time from x-ray CT time-series. The salt layer appears in dark gray while the heterogeneous sedimentary layers appear in light gray. In particular, we show how accurate and ubiquitous the tracking of the two operations of merge and split of the salt layer which deforms under the action of gravity can be performed. 
In Fig~\ref{fig:realistic}, the formation of a salt-walled mini-basin caused by salt deformation and sediment influx is observed. 
From a computational point of view, stochastic filtering that requires a large number of particles is very time and ressources consuming~\cite{ding2016study}.
To speed up computations, we parallelized the original \textit{C++} implementation of~\cite{avenel2014stochastic} using the \textit{Python Dask} library. With 200 particles over 20 observations, running time was reduced from more than 3 hours (sequential) to only 23 minutes (parallel).

\begin{figure}[!htb]
\centering
\includegraphics[width=0.48\textwidth]{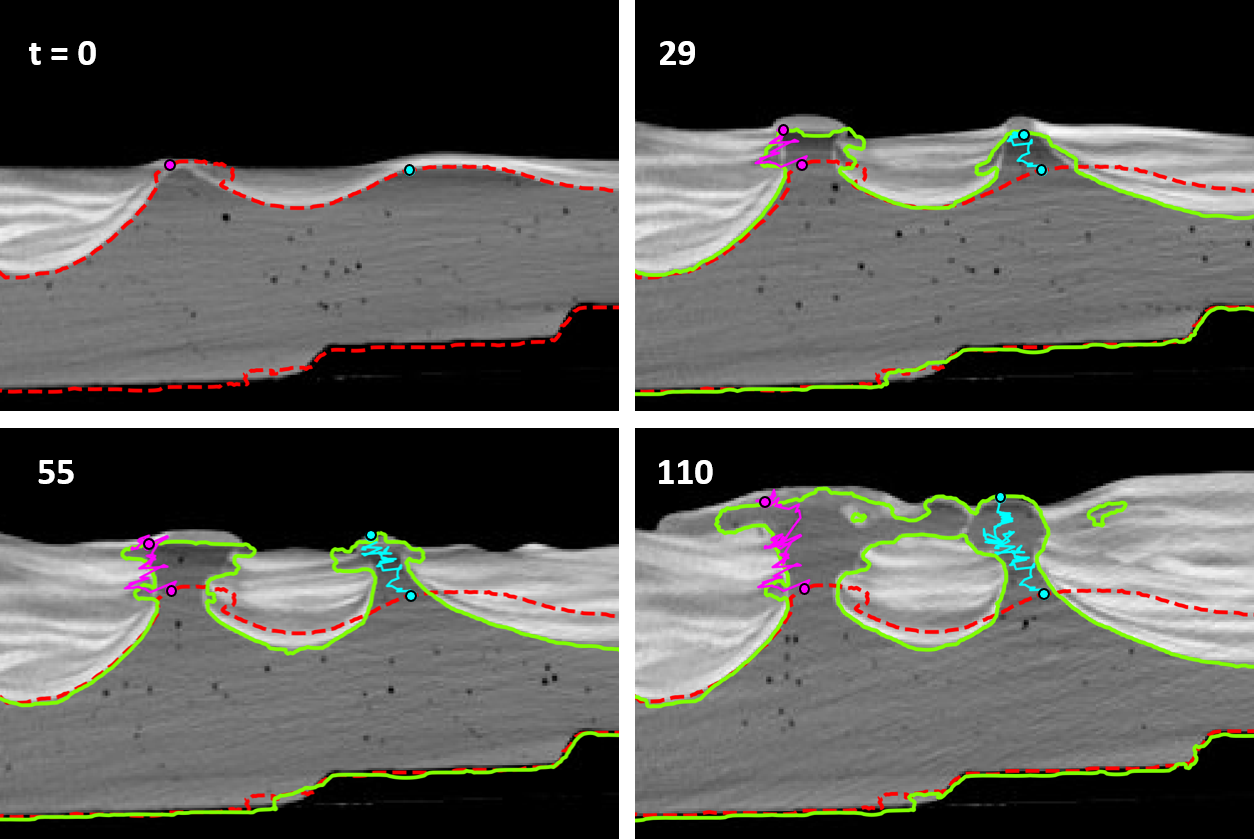}
\caption{\label{fig:realistic} Application to track a realistic salt/rock interface. The forward path of each tracked marker is encoded with a unique color throughout time.}
\end{figure}

\section{Conclusion}
In this paper, we employed an optical-flow-guided particle filter to delineate between the salt and surrounding rocks from CT-image sequences depicting salt flow in sedimentary basins. 
The stochastic approach appears to offer satisfactory results for the segmentation of salt layers undergoing arbitrarily large deformations. Future works will be motivated towards analyzing the sensitivity of model output to noise magnitude (diffusion constants). Also conceivable is the extension of methods from 2D to 3D.

\clearpage
\bibliographystyle{IEEEbib}
\bibliography{refs}

\end{document}